\documentclass[runningheads]{llncs}
\usepackage{graphicx}

\usepackage{tikz}
\usepackage{comment}
\usepackage{amsmath,amssymb} 
\usepackage{color}
\usepackage{bm}
\usepackage[width=122mm,left=12mm,paperwidth=146mm,height=193mm,top=12mm,paperheight=217mm]{geometry}
\usepackage[pagebackref=true,breaklinks=true,letterpaper=true,colorlinks,bookmarks=false]{hyperref}
\usepackage{cuted}
\usepackage{caption} 
\usepackage{subcaption} 
\usepackage{multirow} 
\usepackage{bbm}
\usepackage{booktabs}
\graphicspath{{figure/}}

\begin{document}
\pagestyle{headings}
\mainmatter
\def\ECCVSubNumber{4786}  

\title{On the Burstiness of Faces in Set} 

\author{Jiong Wang*}
\institute{Ningbo University, China}
\def\thefootnote{*}\footnotetext{liubinggunzu@gmail.com}

\maketitle

\begin{abstract}
Burstiness, a phenomenon observed in text and image retrieval, refers to  that particular elements appear more times in a set than a statistically independent model assumes. 
We argue that in the context of set-based face recognition (SFR), burstiness exists widely and degrades the performance in two aspects: Firstly, the bursty faces, where faces with particular attributes 
dominate the training face sets and
lead to poor generalization ability to unconstrained scenarios. Secondly, the bursty faces 
interfere with the similarity comparison in set verification and identification when evaluation. 
To detect the bursty faces in a set, we propose three strategies based on Quickshift++, feature self-similarity, and generalized max-pooling (GMP). 
We apply the burst detection results on training and evaluation stages to enhance the sampling ratios or contributions of the infrequent faces. When evaluation, we additionally propose the quality-aware GMP that enables awareness of the face quality and robustness to the low-quality faces for the original GMP.
We give illustrations and extensive experiments on the SFR benchmarks to demonstrate that burstiness is widespread and suppressing burstiness considerably improves the recognition performance. 
\end{abstract}

\begin{figure}
	\centering
	\begin{subfigure}[t]{0.3\textwidth}
		\centering
		\includegraphics[width=1.23\textwidth]{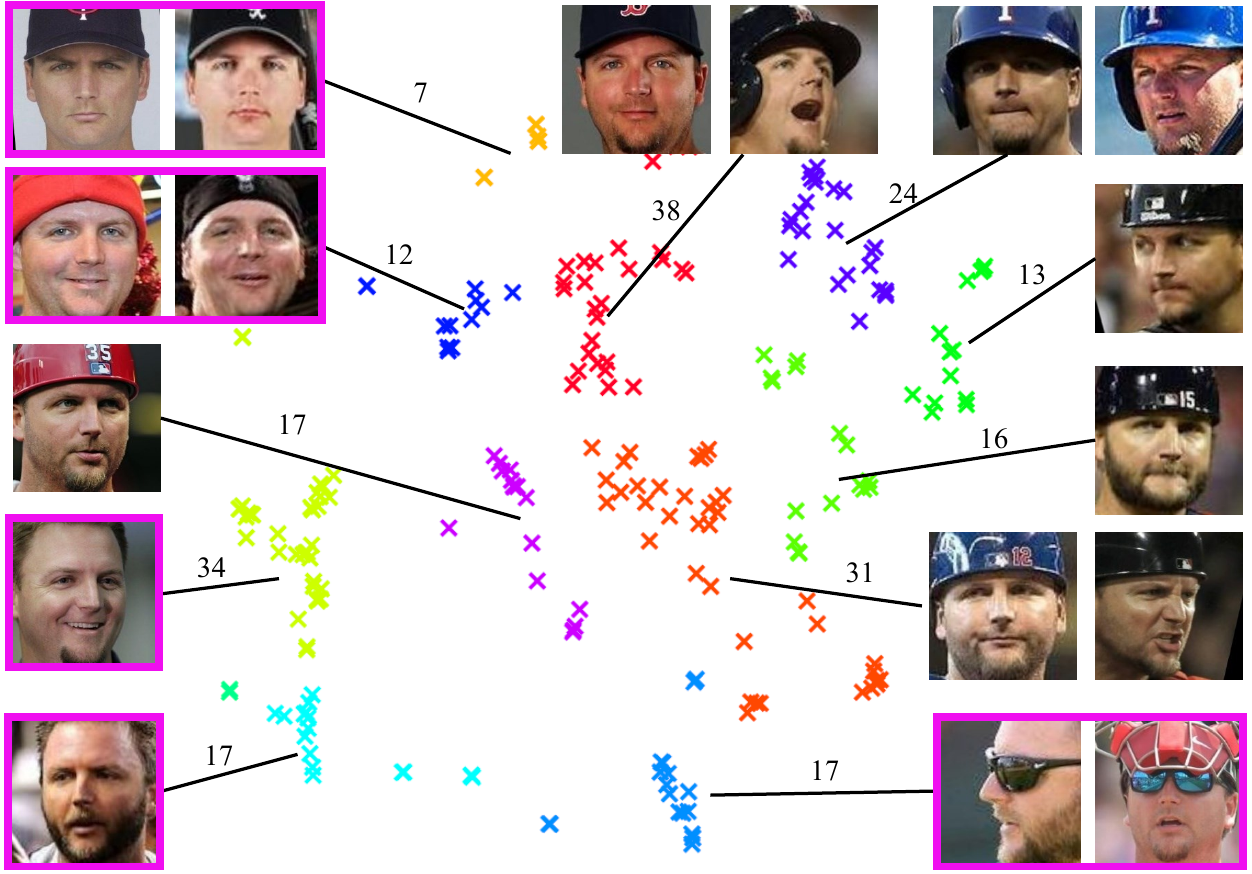}
	\end{subfigure}\hspace{0.7cm}
	\begin{subfigure}[t]{0.3\textwidth}
		\centering
		\includegraphics[width=0.76\textwidth]{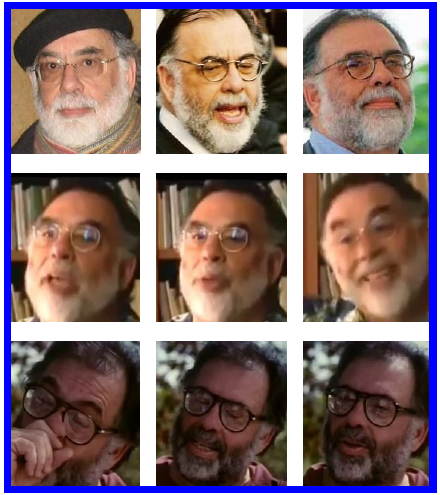}
	\end{subfigure}\hspace{-0.3cm}
	\begin{subfigure}[t]{0.3\textwidth} 
		\centering 
		\includegraphics[width=0.84\textwidth]{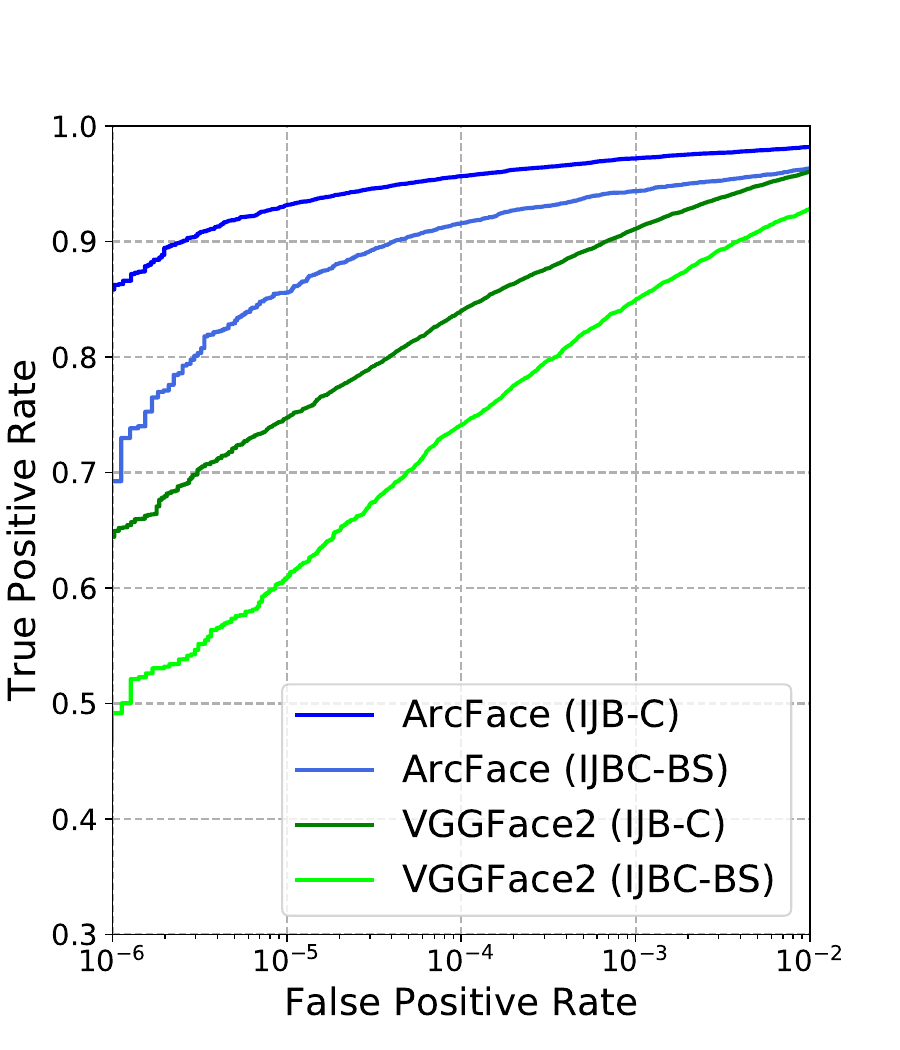}
	\end{subfigure}
	\caption{Illustration of the burstiness phenomenon in training, evaluation datasets, and the ROC curves under the IJBC-BS protocol. (Left) The exemplary training face set is bursting with faces with beard and helmet, while the faces without these attributes are in small numbers (marked with pink border). The face groups are detected by Quickshift++ \cite{jiang2018quickshift++} with cardinalities nearby. 
	(Middle) An evaluation face set may be dominated by video frames with redundancy or faces with similar attributes. (Right) The performance drops dramatically with the IJBC-BS protocol, which focuses on the evaluation of bursty sets in IJB-C \cite{maze2018iarpa}.}
	\label{fig:Motivation}
\end{figure}

\section{Introduction}

\label{sec:intro}

Set-based face recognition (SFR) aims to recognize 
the identity of a set of faces, where each set could contain faces of the same identity from multiple sources (\emph{e.g.}, static images, video frames, or a mixture of both).
SFR has attracted consistent interest in the computer vision community as a growing number of both face images and videos are continually uploaded to the Internet or being captured by the terminal devices. Because of the unconstrained capture environments (in terms of varying poses, illuminations, expressions, and occlusion), it is challenging for a SFR system to get compact and representative set representations.

Facilitated by the success of deep learning, the face recognition (FR) performance is dramatically improved in the past decade. 
The face features extracted from FR model improve the robustness of SFR \cite{schroff2015facenet,taigman2014deepface,cao2018vggface2,deng2019arcface}, where the set-wise representation can be obtained by simply averaging over the face features in set. 
For the Faces in the unconstrained scenario are of quite varying qualities \cite{ferrara2012face,abaza2012quality,hernandez2019faceqnet}, which are associated with feature discriminability.  
Face quality attention is widely studied in existing SFR works \cite{yang2017neural,liu2017quality,xie2018multicolumn,xie2020inducing} to suppress the low-quality faces while strengthening the high-qualities. 
Even though remarkable results are reported, the advances depend mainly on the feature robustness and the elaborate attention modules. The set representations are usually biased to the bursty faces with the sum-aggregation strategy. 
We argue that an ideal set representation should be democratic \cite{murray2016interferences} to all the variations exist in the set, thus can guarantee the generalization to the evaluation in unconstrained scenario. 
As shown in Figure~\ref{fig:Motivation} (left), the great majority of the faces are with helmets,  while the mere portion with hats, sunglasses or none of all. In an exemplary evaluation set (middle), most faces are in older ages while littles are in younger age with distinct attributes. 
When top 5,000 most bursty face sets in IJB-C are selected for evaluation (right, IJBC-BS protocol), the performance drops dramatically compared to whole face sets, which demonstrates the difficulty to recognize bursty face sets.

To detect the bursty faces in face sets, we adopt Quickshift++ \cite{jiang2018quickshift++}, a mode-seeking algorithm, to quantize face groups with similar patterns. 
A group with larger cardinality is deemed as a bursty group.
We also explore two continuous weighting strategies to estimate the burstiness degree of each element. 
One of which is feature self-similarity, based on the assumption that frequent face features are usually close to the feature center of a set. The other is the generalized max-pooling \cite{murray2014generalized,murray2016interferences}, which aims to equalize the similarity between each element feature and set representation. GMP realizes the objective by re-weighting the element features, and we suppose these weights, as a by-product,to indicate the burst degrees of elements.

Having three kinds of detection results, we correspondingly propose three burst-aware sampling strategies in training stage, which enhance the sampling ratios of infrequent faces in a set to enable the model robustness on the unusual conditions.
To adapt the detection results in evaluation stage, we propose corresponding burst-aware aggregation strategies to enhance the contributions of infrequent faces to the set representation. We also propose the quality-aware GMP (QA-GMP), which
combines GMP objective with quality attention scores to overcome the sensitivity of GMP to the low-quality faces.

In summary, the contributions are threefold:
\begin{itemize}
	\item We propose three burstiness detection methods and apply them in the training and evaluation stages to suppress the burstiness in face sets.
	
	\item We use the intermediate product of GMP, the GMP weights, to indicate the burstiness degrees for burst-aware training. We also propose the QA-GMP to guarantee the robustness of GMP to the low-quality faces.
	
	\item We are the first to exhaustively study the burstiness phenomenon in SFR task. The extensive experiments demonstrate the effectiveness of burstiness suppression and we get new state-of-the-art results on the SFR benchmarks.   
	
\end{itemize}

\section{Related Works} \label{sec:related}

\subsection{Set-based Face Recognition}
Set-based face recognition (SFR) has been widely studied and the earlier works represent face set as manifold \cite{lee2003video,arandjelovic2006information,huang2015projection,wang2015discriminant}, convex hull \cite{cevikalp2010face,cevikalp2019discriminatively}
and measure the set distance in corresponding space.
Recently, the well-learned face recognition models \cite{schroff2015facenet,taigman2014deepface,cao2018vggface2,deng2019arcface} demonstrate prominent performance on SFR benchmarks with a simple sum-aggregation strategy.

Face quality attention is a crucial strategy to improve the sum-aggregation and is widely studied in existing SFR works \cite{yang2017neural,liu2017quality,xie2018multicolumn,xie2020inducing} to down-weight the low-quality faces while strengthening the high-qualities. VLAD \cite{jegou2010aggregating,zhong2018ghostvlad} is also adopted to aggregate the face features to a compact representation.

However, these works ignore the burstiness in the face sets, which harms the generalization ability of face encoder in the training stage, and causes the set representation to be dominated by the frequent features. 
We propose to suppress burstiness with burst-aware sampling and aggregation strategies separately in the training and evaluation stage. 

\subsection{Visual Burstiness Analysis}
Inspired the burstiness analysis in text recognition \cite{church1995poisson,madsen2005modeling,he2007using}, the burstiness phenomenon was brought to attention in the image retrieval works \cite{torii2013visual,jegou2009burstiness,shi2015early}, where the repetitive structures in natural images lead to burstiness and corrupt the visual similarity measurement.
The power normalization \cite{perronnin2010improving,koniusz2018deeper} and intra-normalization \cite{arandjelovic2013all} are also considered to discount the contribution of visual bursts. 

Derived from similar inspiration to reduce burstiness for aggregation, J{\'e}gou \emph{et al.} \cite{jegou2014triangulation} and Murray \emph{et al.} \cite{murray2014generalized} separately propose the democratic aggregation and the generalized max-pooling. \cite{jegou2014triangulation} adopts a modified sink-horn algorithm to iteratively calculate the democratic weights, while \cite{murray2014generalized} calculates the GMP weights by solving the linea-regression problem. Comparatively, GMP performs better as demonstrated in \cite{murray2016interferences}.

We use the intermediate product of GMP, the GMP weights, to indicate the burstiness degrees of samples in set. These weights are used in the proposed burst-aware sampling and aggregation methods.
Moreover, we propose the QA-GMP to guarantee robustness to the low-quality faces, which consistently improves GMP on SFR task.

\subsection{Bias in Face Recognition}

Face recognition models are known to be discriminative on particular demographic groups (e.g., races, genders, ages) dominating the training set, while performing worse on the rare groups \cite{howard2019effect,gong2021mitigating}. To mitigate these biases, 
the re-weighting and re-sampling strategies are studied \cite{amini2019uncovering,morales2020sensitivenets,more2016survey,chawla2002smote,akbani2004applying} in the training process. 
The demographic-adaptive margin loss \cite{wang2020mitigating} and demographic-adaptive convolution kernels \cite{gong2021mitigating} are recently proposed to adaptively optimize different groups for fair face features.

Different from the aforementioned works which consider the inter-identity bias in the whole-dataset level, we consider the intra-identity burstiness in a face set. Moreover, we consider both the burst-aware training and evaluation in SFR task.

\section{Method} \label{sec:method}

In this section, we first introduce the proposed set-based face recognition (SFR) pipeline (Section \ref{sec:Preliminary}), then detail the burstiness detection methods (Section \ref{sec:bdetection}) and the adaptation of detection results on training  (Section \ref{sec:btrain}) and evaluation (Section \ref{sec:beval}) stages. We finally describe the IJBC-BS protocol (Section \ref{sec:ijbcbs}). 

\subsection{Preliminary} \label{sec:Preliminary}
\begin{figure}[t]
	\begin{center}
		\includegraphics[scale=0.45]{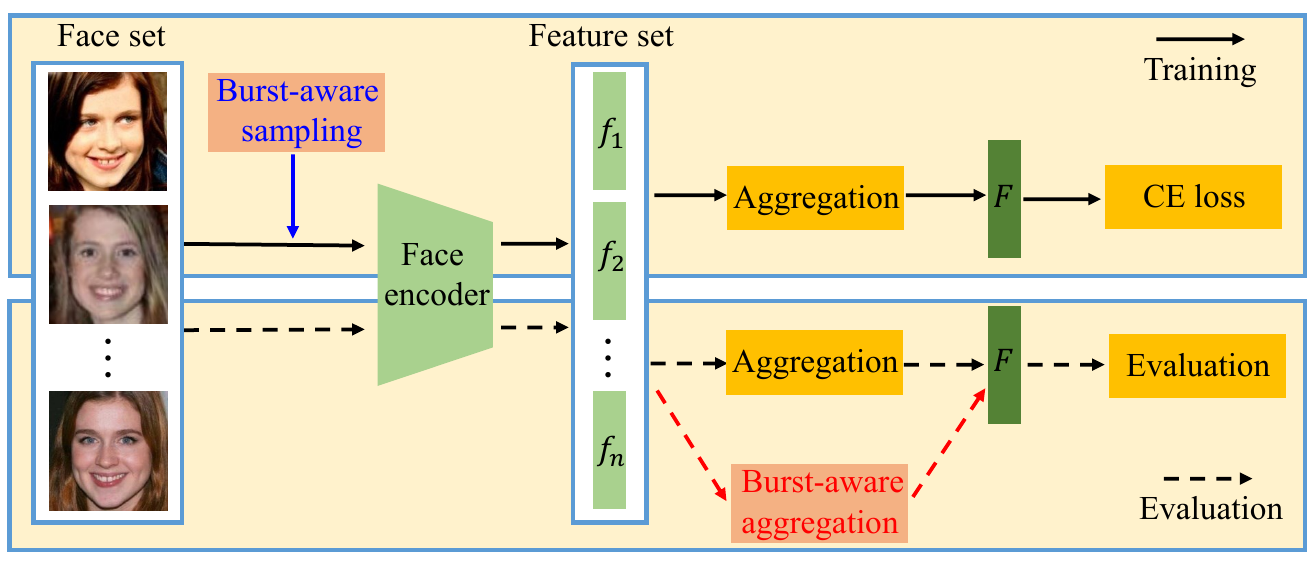}
	\end{center}
	\vspace{-0.2cm}
	\caption{The pipeline to get the set representation. We improve the vanilla pipeline with burst-aware (instance) sampling and burst-aware aggregation in training and evaluation stages, separately.
	}
	\label{fig:pipeline}
	\vspace{-0.2cm}
\end{figure}

A vanilla pipeline to get the set representation is illustrated in Figure~\ref{fig:pipeline}, where the faces in a set are passed to the face encoder and the resulted face features are aggregated in the aggregation module to the set representation. 
The aggregation module consists of 
an attention block to weigh each feature and the sum-aggregation to get the set representation. For SFR task, the pre-trained face encoder is assumed to be available and usually the pre-trained dataset is used for fine-tuning \cite{xie2018multicolumn,zhong2018ghostvlad}. So we define the fine-tuning process as training. 

Supposing a face set $\mathcal{X} = \{x_1, x_2, ..., x_{n}\}$ consists of $n$ faces with same identity. 
$\mathcal{X}$ is passed to the face encoder and the feature set $X = \{f_1, f_2, ..., f_n\}$ is obtained with shape of $n \times d$.
In the aggregation module a simple attention block \cite{yang2017neural,xie2018multicolumn} re-weights all the faces, 
where a $d$ dimensional query vector $q$ evaluates the face features and gets corresponding attention scores. The sigmoid function is used to scale the numerical values and get the attention scores. 
Finally, sum aggregation with attention weights gets the $d$ dimensional set representation $F$:
\begin{equation}
F = \sum\nolimits_{i=1}^{n} \alpha_{i}^{att} f_i, \  \  \alpha_{i}^{att} = \bm{\sigma} (q f_i^T).
\end{equation}

In each training epoch, all the identities are sampled and each identity corresponds to one training instance (face set), which consists of $n_t$ randomly sampled faces from a corresponding face set. 
The set representation in Figure~\ref{fig:pipeline} is used for identity classification with cross-entropy loss, where the face encoder and aggregation module are optimized to get discriminative set representations for evaluation. 
To suppress burstiness in face sets, we separately propose the burst-aware instance sampling and burst-aware aggregation in training and evaluation stages.

\subsection{Burstiness Detection} \label{sec:bdetection}
As noted above, the training instances are randomly sampled from the face sets. The frequent faces dominating the face set naturally dominate the instance. A similar phenomenon exists in the evaluation stage, where the set representation is biased to the frequent face features with sum aggregation.
Before suppressing the bursty faces, we first describe the strategies to detect the bursty faces in set.

An intuitive method for burstiness detection is to quantize the groups with similar patterns \cite{shi2015early}, so we resort to a mode-seeking method: Quickshift++ (Q-shift). We also explore two continuous weighting strategies based on the feature self-similarity (S-sim) and generalized max-pooling (GMP).

\noindent\textbf{Quick-shift++} \cite{jiang2018quickshift++} is a fast and stable mode-seeking method to discover group genes with similar expression patterns.
Compared to K-Means, Q-shift can discover clusters of arbitrary shape and don't require the prior about cluster number. 

A face set is partitioned to $n_q$ groups by Q-shift, formulated as $\mathcal{X} = \{G_1, G_2, \\..., G_{n_q}\}$. 
An example of the clustering results is shown in Figure~\ref{fig:Motivation} (left), the faces in the same group usually exhibit similar appearance. 
A group with relatively large cardinality is supposed to be a bursty group, which will be suppressed in the training and evaluation stages.

\noindent\textbf{Self-similarity}   
is based on the prior \cite{amini2019uncovering} that the frequent features dominating the set are usually close to the feature center while the infrequent features are far from the center. 
Having the feature set $X$ with shape of $n \times d$, we can get the $n \times n$ self-similarity (Gram) matrix $K$.  Then we average the column of $K$ to get the self-similarities $S$ of the elements in set. 
\begin{equation} \label{eq:gram}
S = \frac{K \bm{1}_n}{n} , K = X X^T.  
\end{equation}
The self-similarities range from -1 to 1 and the elements with higher self-similarities are supposed to be bursty. 

\noindent\textbf{Generalized Max Pooling (GMP)}
\cite{murray2014generalized,murray2016interferences} is devised to equalize the similarity between each element and the aggregated set representation, which is realized by re-weighting the per-element feature. The GMP weights, as an intermediate product, are supposed to reflect the burst degrees of samples in this paper.

Having the  feature set $X = \{f_1, f_2, ..., f_n\}$, let $F_{gmp}$ denotes the aggregated GMP representation, which is expected to be equally similar to the frequent and rare faces, as formulated:
\begin{equation} \label{eqn:gmp1}
f_{i} F_{gmp}^T = \bm{1}, for \ i = 1, 2, ..., n,   
\end{equation}
where $\bm{1}$ can be replaced by an arbitrary constant because the final representation is typically $L_2$-normalized. In matrix form, Equation~\ref{eqn:gmp1} can be rewritten as: 

\begin{equation} \label{eqn:gmp2}
X F_{gmp}^T = \bm{1}_n,   
\end{equation}
which is a linear system of $n$ equations with $d$ unknowns. In general, this system might not have a solution when $d < n$ or might have an infinite number of solutions when $d > n$. Therefore, we turn Equation~\ref{eqn:gmp2} into a least-squares regression problem:
\begin{equation} \label{eqn:gmp3}
F_{\text{gmp}} = \mathop{\arg\min}_{F} 
\lVert X F_{gmp}^T - \bm{1}_n \rVert_2^2 +
\lambda \lVert F\rVert_2^2\;,
\end{equation} 
where $\lambda$ is the regularization term that stabilizes the solution.

Equation~\ref{eqn:gmp3} is the primal formulation of the GMP which does not rely on the explicit computation of a set of weights. 
Murray \emph{et al.} \cite{murray2014generalized} solves the dual formulation of GMP to explicitly compute their weights and  in this way $F_{gmp} = X^T \alpha$. By inserting  it into Equation~\ref{eqn:gmp3}, we obtain
\begin{equation}   \label{eqn:gmp4}
\begin{split} 
F_{gmp} = {}& \mathop{\arg\min}_{\bm{\alpha}} \lVert XX^T \bm{\alpha} - \bm{1}_n \rVert^2 + \lambda \lVert X^T\bm{\alpha} \rVert^2
\\
={}& 
\mathop{\arg\min}_{\bm{\alpha}} \lVert K \bm{\alpha} - \bm{1}_n \rVert^2 + \lambda
\bm{\alpha}^T K \bm{\alpha} \;,
\end{split}
\end{equation}
where $K$ is the gram matrix. Equation~\ref{eqn:gmp4} yields the closed-form solution:
\begin{align} \label{eqn:gmp5}
\bm{\alpha} = (K + \lambda \bm{I}_n)^{-1} \bm{1}_n.
\end{align}
In this way, the GMP weights $\alpha$ can be used to indicate the burstiness of elements in set and the frequent elements usually get lower weights.

\begin{figure}[t]
	\centerline{
		\subcaptionbox{\small S-sim weight \emph{vs.} loss}{
			\includegraphics[scale = 0.28]{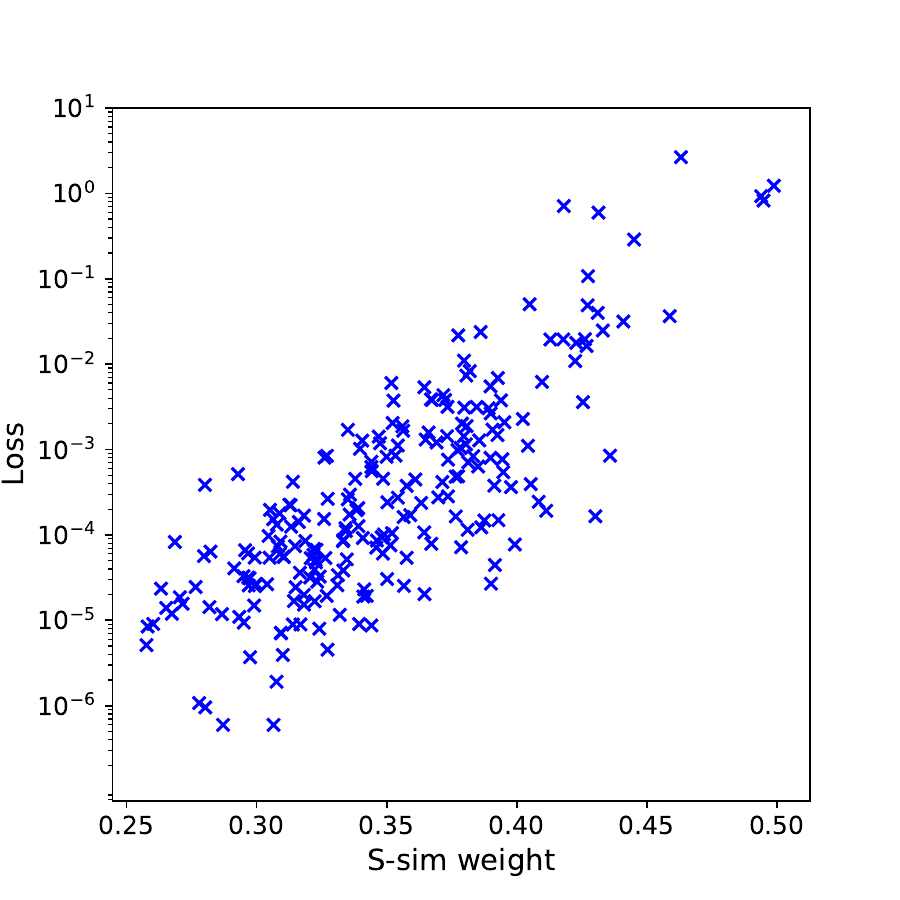}
		} \hspace{-0.5cm}
		\subcaptionbox{\small GMP weight \emph{vs.} loss}{
			\includegraphics[scale = 0.28]{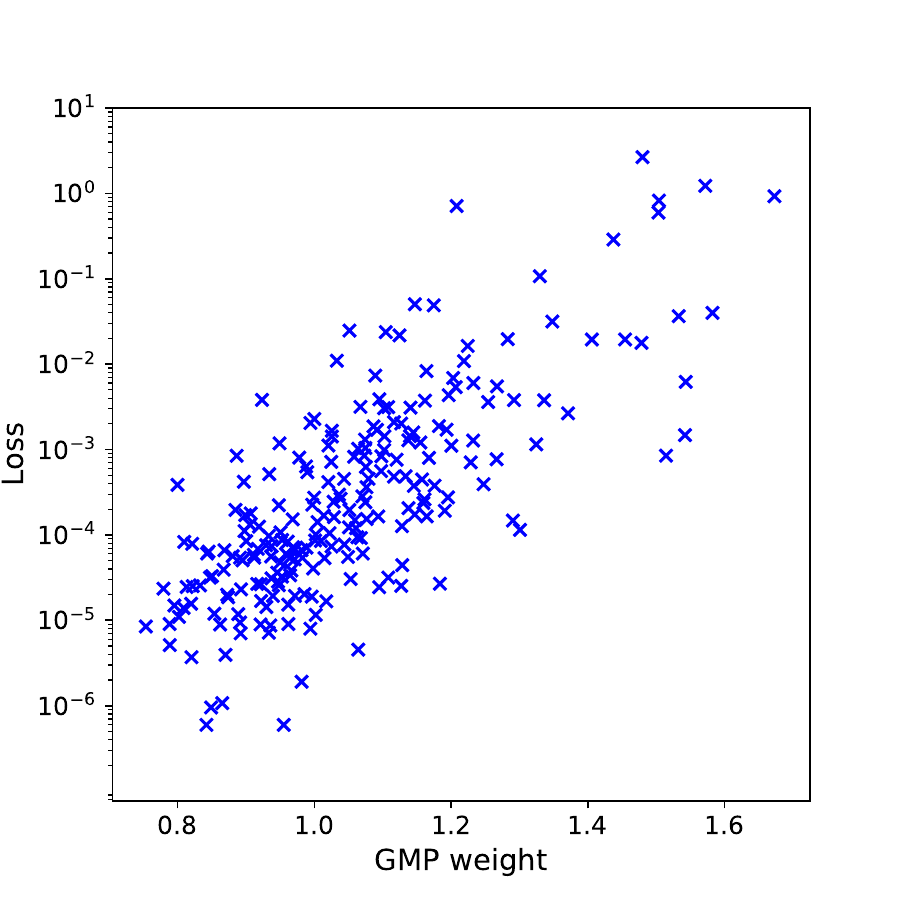}
		}\hspace{-0.1cm}
		\subcaptionbox{\small Sampling weights}{
			\includegraphics[scale = 0.18]{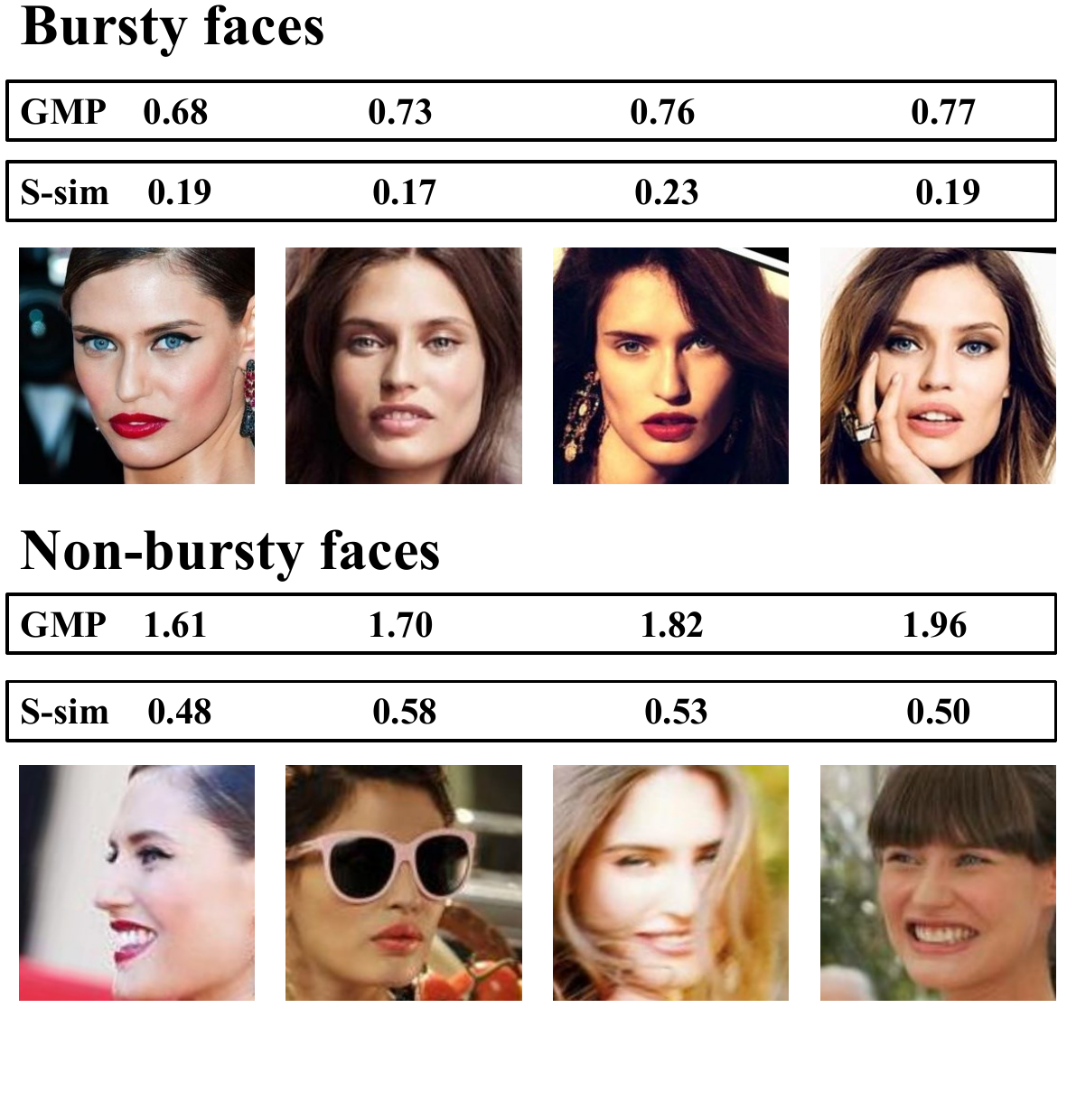}
		}
	}  \vspace{-0.2cm}
	\caption{Illustration of the sampling weights and corresponding losses of samples (a, b). Exemplary faces with sampling weights (c) in a training face set. 
	}
	\label{fig:burst_train}
	\vspace{-0.2cm}
\end{figure}

\subsection{Burst-aware Instance Sampling} \label{sec:btrain}
In conventional SFR pipeline, a training instance with $n_t$ faces is randomly sampled with uniform probability for all the faces in a set. 
Having the burstiness detection results in section \ref{sec:bdetection}, we propose the burst-aware sampling strategy with aim to suppress the sampling ratios of bursty faces while highlighting the infrequent faces for the training instance.

\noindent\textbf{Q-shift}. Given the detected face groups $\{G_1, G_2, ..., G_{n_q}\}$ from Q-shift, with cardinalities $\{c_1, c_2, ..., c_{n_q}\}$. We first select the $n_t$ groups, then randomly sample one face in each selected group to construct the training instance. 
To tackle the extreme situation that the cardinality contrast among groups are 
significant (e.g., $100 : 1$), we correlate the sampling weight of a group to the cardinality as follow:
\begin{equation}   
\alpha_i^{q} = c_i^{\lambda_1},
\end{equation}
where $\lambda_1$ is the scaling factor and it is equivalent to vanilla random sampling when $\lambda_1 = 1$, and $\lambda_1 = 0$ selects groups with uniform probability.

\noindent\textbf{S-sim and GMP}. 
The burstiness weights of self-similarity and GMP are in different numerical ranges and contrast degrees.
To get reasonable range for S-sim, we transform $S$ in Equation~\ref{eq:gram} to the sampling weight as follow: 
\begin{equation} \label{eq:ssimweight}
\alpha_i^{sim} = (1 - S_i)^{\lambda_2}, 
\end{equation}
where $\lambda_2$ is a parameter for scaling. It is intuitive that the bursty features with larger self-similarities get lower sampling probabilities. 
In a similar way, we transform the GMP weight in Equation~\ref{eqn:gmp5} to the sampling weight as follow: 
\begin{equation} \label{eq:gmpweight}
\alpha_i^{gmp} = e^{\lambda_3 \alpha_i},
\end{equation}
The aforementioned sampling weights are finally normalized within a set to the sampling probabilities.

 We found the sampling weights generated by the self-similarity and GMP are correlative to the recognition hardness. 
The losses (calculated from the pre-trained VGGFace2 model \cite{cao2018vggface2}) and the bursty weights of a face set are plotted
in Figure~\ref{fig:burst_train} (a, b). The frequent faces with lower S-sim or GMP weights are usually well-learned in the pre-trained face encoder with lower losses, while the infrequent faces are on the opposite. It can be observed in Figure~\ref{fig:burst_train} (c) that the unusual faces are usually in extreme conditions of poses, occlusions, illumination or age, which are difficult to identify. Thus burst-aware sampling can be seen as a hard-example mining strategy.
But compared to former hard example learning works \cite{shrivastava2016training,xie2018comparator}, it is unsupervised, without accessing the classifier or increasing the training overhead. 

\subsection{Burst-aware Aggregation} \label{sec:beval}
The proposed  burst-aware aggregation strategy aims to suppress the contribution of bursty faces in the aggregation of face features. For \textbf{Q-shift} with $n_q$ detected face groups in a set, we propose a two-stage aggregation method, which first averages the features intra-group to group feature, performs $L_2$-normalization, then aggregates all the group features to the set representation. 

For \textbf{self-similarity} and \textbf{GMP}, the solution is to apply weighted aggregation with the bursty weights. Together with the attention scores,  we get the set representation as formulated: 
\begin{equation} 
F_{gmp} = \sum\nolimits_{i=1}^{n}  \alpha_{i}^{att} \alpha_i^{sim} f_i.
\end{equation} 
This formulation is also applied to the aggregation of GMP or QA-GMP weights.

\begin{figure}[t]
	\centering
	\begin{subfigure}[t]{0.5\textwidth}
		\centering
		\includegraphics[scale = 0.32]{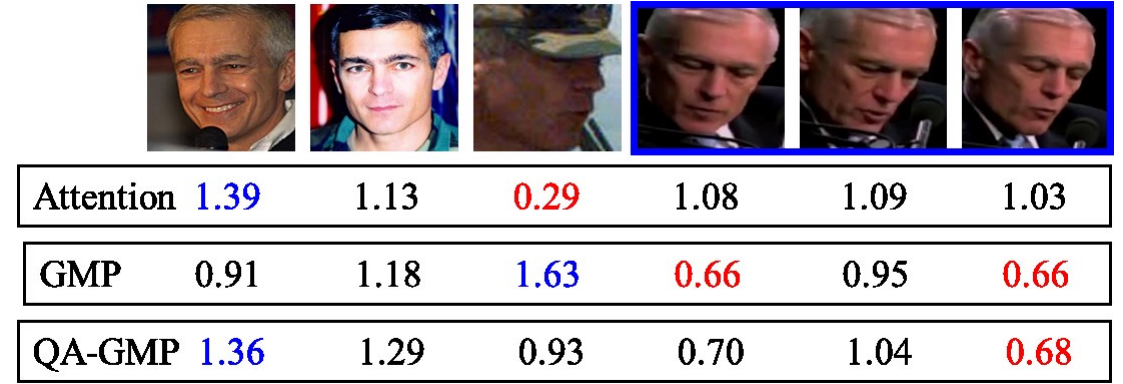}
	\end{subfigure}%
	\begin{subfigure}[t]{0.5\textwidth}
		\centering
		\includegraphics[scale = 0.32]{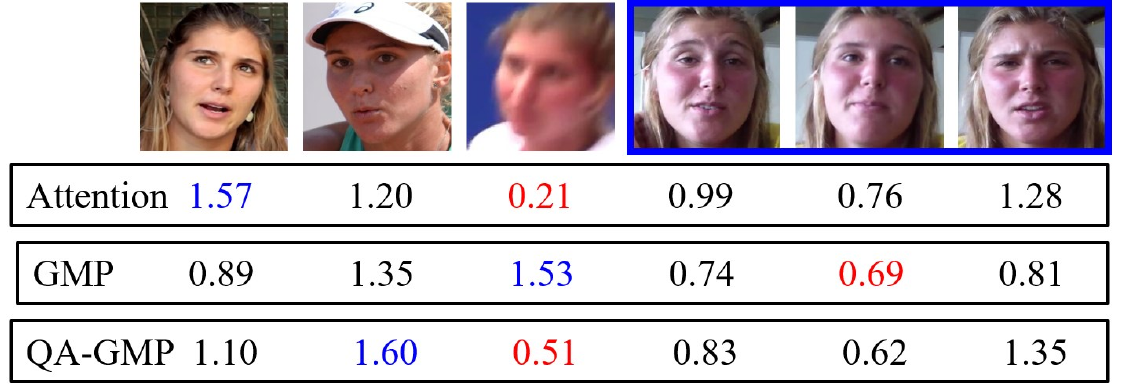}
	\end{subfigure}%
	\caption{Illustration of the attention, GMP, and QA-GMP scores. The scores are normalized for comparison.  The highest and lowest values are highlighted in blue and red.}
	\label{fig:burst_score}
\end{figure}

\noindent\textbf{Quality-aware generalized max-pooling.} 
GMP works on the assumption that all the elements in a set are informative, while it is  sensitive to the low-quality faces. 
We found the GMP usually generates larger weights to the low-quality faces, whose features are not well-learned and thus be considered as infrequent. 
As shown in Figure~\ref{fig:burst_score}, the attention weights are quality-aware, while the GMP weights are frequency-aware. In both examples, 
the third face with low quality gets lowest quality score but highest GMP score. However, giving higher weights to low-quality faces  harms the discriminability of set representation.

To endow the GMP with awareness of face quality, we propose the quality-aware GMP (QA-GMP) to combine GMP with attention scores. QA-GMP regularizes the GMP with attention scores and rewrites the objective (Equation~\ref{eqn:gmp2}) as follow:
\begin{equation} \label{eqn:qagmp1}
X^T F_{gmp} = \bm{1}_n + \bm{\lambda_4 \alpha^{att}},  
\end{equation}
and the corresponding solution (Equation~\ref{eqn:gmp5}) is rewritten as:
\begin{equation}
\bm{\alpha^{qag}} = (K + \lambda \bm{I}_n)^{-1} (\bm{1}_n + \bm{\lambda_4 \alpha^{att}}),
\label{eq:}
\end{equation}
where $\lambda_4$ is a factor to balance the impact of quality and frequency.

With this modification, QA-GMP is aware of the face quality with $\alpha^{att}$ as regularization, when equalizing the similarity between each element and the set representation.  
Thus QA-GMP is robust to low-quality faces and consistently improves GMP as demonstrated in Figure~\ref{fig:burst_score} and Section~\ref{sec:experiment}. 

\noindent\textbf{Discussion.} 
As a clustering algorithm, Q-shift may get non-robust results on the small sets with lower cardinalities. Hence we only apply Q-shift on training datasets and the evaluation of YTF dataset \cite{wolf2011face}, not on IJB datasets \cite{whitelam2017iarpa,maze2018iarpa}.  We found Q-shift takes effect on the $n < d$ (cardinality smaller than dimensionality) scenario for both training and evaluation stages, which will be detailed in the supplementary.

Even though S-sim and GMP consist of differentiable operations, we found applying burst-aware aggregation in training stage gets worse results. So burst-aware aggregation is only applied in evaluation stage.

\subsection{IJBC-BS Subset} \label{sec:ijbcbs}
The 1:1 verification protocol on IJB-C dataset \cite{maze2018iarpa} evaluates the pair-wise verification of 23,124 face sets with static face images and video frames. 
Even though remarkable performance is reported on the whole dataset, we found the performance on bursty face sets is fabulously worse. 

We define the burst degree of a set by the mean value of the self-similarity matrix (Equation~\ref{eq:gram}). With this indicator, 5,000 face sets with top burst degrees are selected from IJB-C and constitute the IJBC-BS subset. 
As shown in Figure~\ref{fig:Motivation} (right), the performance drops dramatically on IJBC-BS compared to IJB-C. Detailed statistics on the IJB-BS subset are given in the supplementary.

\section{Experiments} \label{sec:experiment}

\subsection{Datasets}
We adopt the VGGFace2 \cite{cao2018vggface2} (SENet-50 \cite{hu2018squeeze}) or the ArcFace \cite{deng2019arcface} (ResNet-100 \cite{he2016deep}) model as face encoder, which is separately pre-trained on the VGGFace2 \cite{cao2018vggface2} or a cleaned version  of MS1M \cite{guo2016ms} dataset. The face encoders are fine-tuned for set-based face recognition on the same dataset with the pre-training model.

The unconstrained face recognition benchmarks, IJB-B \cite{whitelam2017iarpa} and IJB-C \cite{maze2018iarpa}, are used for evaluation. IJB-B and IJB-C are mix-sourced (video frames and images) benchmarks for set-based verification (1:1) and identification (1:N). We also have experiments on the Youtube Face (YTF) dataset \cite{wolf2011face} which is video-sourced and the set-based verification performance is evaluated. The performance on IJBC-BS, a subset of IJBC, is also reported.

\subsection{Implementation Details}

The MTCNN algorithm \cite{zhang2016joint} is adopted for face detection and alignment. Following previous works \cite{cao2018vggface2,deng2019arcface}, faces are detected, cropped, without aligned, and resized to $224 \times 224 \times 3$ as inputs of VGGFace2 model. Faces are detected, cropped, aligned, and resized to $112 \times 112 \times 3$ for ArcFace model.
A dropout layer with dropout ratio 0.1 is adopted following the face encoder.  For VGGFace2 model, we additionally apply a fully-connected layer to reduce the feature dimensionality to 256 before aggregation.
For fine-tuning, a training instance consists of 15 faces with different sampling strategies. We froze the parameters of face encoder and only the aggregation module is optimized with the RMSprop optimizer with batch size of 128, learning rate is 0.001, momentum is 0.9. We set the parameters in Section~\ref{sec:method} $\lambda = 1$, $\lambda_1 = 0.5$, $\lambda_2 = 2$, $\lambda_3 = 10$ and $\lambda_4 = 5$.

\begin{table}[t]
	\centering
	\begin{tabular}{l|c|c|c|c}
		\toprule
		\multirow{2}*{\textbf{Method}} & \multicolumn{2}{c|}{\textbf{IJB-B 1:1 TAR (\%)}} & \multicolumn{2}{c}{\textbf{IJB-C 1:1 TAR (\%)}}\\
		\cline{2-5} 
		~ & FAR=1e-6 & FAR=1e-5 & FAR=1e-6 & FAR=1e-5 \\
		\hline
		VGGFace2 \cite{cao2018vggface2} &36.56 & 67.63&64.94 &74.71  \\
		+ Vanilla  &   34.51 & 72.82  & 65.79& 78.73  \\
		+ \textbf{Q-shift (T)} & 38.50  & 74.60 &  68.81 & 81.11  \\
		+ \textbf{S-sim (T)} & \textbf{39.93} & 75.00 & 69.58 & \textbf{81.54}      \\
		+ \textbf{GMP (T)} &  37.92 & \textbf{75.98}  &  \textbf{71.59} & 81.43  \\
		\hline
		ArcFace \cite{deng2019arcface} &38.28 &89.33&86.25 &93.15  \\
		+ Vanilla &  45.13 & 89.38  & 88.55 &93.84  \\  
		+ \textbf{Q-shift (T)} &  45.97 &89.40 & 89.32&93.94\\ 
		+ \textbf{S-sim (T)}&  46.33 &\textbf{ 89.51} & 89.41& 94.03\\
		+ \textbf{GMP (T)} & \textbf{47.50} & 89.35 & \textbf{89.85}&\textbf{94.06}   
		\\\bottomrule
	\end{tabular}
	\caption{Ablation of burst-aware training on IJB-B and IJB-C 1:1 verification protocols. The best results are highlighted in bold.}
	\label{tab:burst_train}
	\vspace{-0.5cm}
\end{table}

\subsection{Ablation Study}
In this subsection, we give exhaustive ablation studies to demonstrate the effectiveness of burstiness suppression in the training and evaluation stage.

\noindent\textbf{Burst-aware instance sampling.} In Table~\ref{tab:burst_train}, we compare the vanilla training pipeline with the proposed three sampling strategies. It can be seen that vanilla training moderately improves the performance of VGGFace2 and ArcFace models on the IJB-B and IJB-C benchmarks. The proposed sampling strategies: Q-shift (T), S-sim (T), and GMP (T) consistently surpass the vanilla training by suppressing the bursty faces and highlighting the infrequent faces. The three strategies considerably improve the original backbones by large margins. 
We can observe S-sim and GMP outperform the Q-shift, and GMP usually performs best.

\begin{table}[t]
	\centering
	\begin{tabular}{l|c|c|c|c|c}
		\toprule
		\multirow{2}*{\small\textbf{Method}} &\multirow{2}*{\small\textbf{Train}} & \multicolumn{2}{c|}{\textbf{IJB-B 1:1 TAR (\%)}} & \multicolumn{2}{c}{\textbf{IJB-C 1:1 TAR (\%)}}\\
		\cline{3-6} 
		~ & & FAR=1e-6 & FAR=1e-5 & FAR=1e-6 & FAR=1e-5 \\
		\hline
		GMP (T) &\multirow{4}*{\small VGGFace2} & 37.92 & 73.50  &  \textbf{71.59} & 81.43\\
		+ S-sim (E)& &41.04 &76.23 & 71.19 &81.64\\
		+ GMP (E)& &\textbf{41.44} & 76.30  & 71.00 & 81.70 \\
		+ QA-GMP (E)&&41.15 &\textbf{76.77} & 71.34&\textbf{82.09}\\
		\hline
		 GMP (T) &\multirow{4}*{\small MS1M}&47.50 & 89.35 & 89.85&94.06\\ 
		+ S-sim (E)& &47.59 & 89.45 &89.66&94.01 \\
		+ GMP (E)& &48.90 & 89.39 &89.83& 94.04 \\
		+ QA-GMP (E)& &\textbf{49.04} &\textbf{89.62}& \textbf{90.87}&\textbf{94.41}
		\\\bottomrule
	\end{tabular}
	\caption{Ablation of burst-aware aggregation on IJB-B and IJB-C 1:1 verification protocols with GMP training strategy.}
	\label{tab:burst_agg1}
	\vspace{-0.5cm}
\end{table}

\noindent\textbf{Burst-aware aggregation.} 
We apply the proposed burst-aware aggregation (E) strategies on the best-performed GMP (T) model and present the result comparison with vanilla sum aggregation in Table~\ref{tab:burst_agg1}.  
It can be observed that suppressing bursty faces with the proposed three methods improves the vanilla sum-aggregation.  
Specifically, the performance advances are not significant for S-sim and GMP aggregation. We suppose the reason is they consider only face frequency but ignore quality, which will be interfered with the low-quality faces and gets higher aggregation weights for them.
Comparatively, the QA-GMP is aware of both image quality and frequency, thus is more robust than S-sim and GMP on the unconstrained scenario.

\begin{table}[t]
	\begin{minipage}[t]{0.5\textwidth}
		\centering
		\begin{tabular}{l|c|c}
			\toprule
			\multirow{2}*{\textbf{Method}} & \multicolumn{2}{c}{\textbf{1:1 TAR}} \\
			\cline{2-3}
			~ & FAR=1e-6 & 1e-5 \\
			\hline
			Vanilla & 65.79 & 78.73 \\
			\hline
			+ P-norm \cite{perronnin2010improving} &65.14 & 78.50 \\ 
			+ DA \cite{jegou2014triangulation} & 65.86 & 79.12   \\
			\hline
			+ S-sim (E) & 66.12 & 79.09 \\
			+ GMP (E) & 65.97 & 79.16\\
			+ QA-GMP (E) & \textbf{67.70} & \textbf{79.75}  \\
			\bottomrule
		\end{tabular}
		\caption{Comparison of different burstiness suppression methods on IJB-C.}
		\label{tab:burst_comp}
	\end{minipage}
	\begin{minipage}[t]{0.5\textwidth}
		\centering
		\begin{tabular}{l|c|c}
			\toprule
			\multirow{2}*{\textbf{Method}} & \multicolumn{2}{c}{\textbf{1:1 TAR}} \\
			\cline{2-3}
			~ & FAR=1e-6 & 1e-5 \\
			\hline
			FaceQNet \cite{hernandez2019faceqnet} &66.86 & 79.83 \\ 
			\textbf{+ QA-GMP} &   \textbf{68.75} & \textbf{80.65} \\
			\hline
			NAN \cite{yang2017neural} & 67.59 & 80.17  \\
			\textbf{+ QA-GMP} &  \textbf{69.17} & \textbf{81.21} \\
			\hline
			MCN \cite{xie2018multicolumn} & 70.09 & 81.46  \\
			\textbf{+ QA-GMP} & \textbf{71.28} & \textbf{82.06} \\
			\bottomrule
		\end{tabular}
		\caption{Performance of QA-GMP with different face quality scores on IJB-C.}
		\label{tab:face_quality}
	\end{minipage}
	
\end{table}

\noindent\textbf{Comparison with existing burstiness suppression methods.}
Even though burstiness is not well-studied in deep learning models, the power normalization (P-norm) \cite{perronnin2010improving} and democratic aggregation (DA) \cite{jegou2014triangulation} can be applied to suppress the burstiness. We compare these two methods with the proposed three strategies on the vanilla attention block based on VGGFace2 model in Table~\ref{tab:burst_comp}. It can be seen that all the methods except QA-GMP barely or slightly take effect. We suppose the reason is they are interfered with the low-quality faces in the unconstrained scenario, while the proposed QA-GMP is robust to these interferences.

\begin{table*}[t]
	\centering
	\begin{tabular}{l|c|c|c|c|c|c|c}
		\toprule
		\multirow{2}*{\textbf{Method}} &\multirow{2}*{Train set} & \multicolumn{3}{c|}{\textbf{IJB-B 1:1 TAR}} & \multicolumn{3}{c}{\textbf{IJB-C 1:1 TAR}} \\
		\cline{3-8} 
		~ &&FAR=1e-6& 1e-5 & 1e-4 & FAR=1e-6& 1e-5 & 1e-4 \\
		\hline
		VGGFace2 \cite{cao2018vggface2} &\multirow{7}*{VGGFace2}& -&67.1 & 80.0 & 61.7 &75.3&85.2\\ 
		MCN \cite{xie2018multicolumn}& && 70.8 & 83.1 & 64.9 &77.5&86.7  \\  
		DCN \cite{xie2018comparator} & &-&- & 84.9 & -&- & 88.5 \\
		GhostVLAD \cite{zhong2018ghostvlad} && - &76.2 & 86.3 &-&-&-   \\
		PCNet \cite{xie2020inducing} & &- &- &- &69.5 &80.0&\textbf{89.0} \\
		\textbf{GMP (T)}& &37.92&75.98&86.26&\textbf{71.59}&81.43&88.79 \\
		\textbf{+ QA-GMP (E)}& &\textbf{41.15}&\textbf{76.77}&\textbf{86.37}&71.34&\textbf{82.09}&\textbf{88.97}\\
		\hline
		PFE \cite{shi2019probabilistic} & \multirow{4}*{MS1M} &-&-&-&-& 89.64& 93.25 \\ 
		ArcFace \cite{deng2019arcface} & &38.28& 89.33&\textbf{94.25}& 86.25&93.15&95.65\\   	
		\textbf{GMP (T)}& &47.50&89.35&93.80&89.85&94.06&95.95\\
		\textbf{+ QA-GMP (E)}&&\textbf{49.04} &\textbf{89.62}&93.86& \textbf{90.87}&\textbf{94.41}&\textbf{96.02}
		\\\bottomrule
	\end{tabular}
	\caption{Compared with state-of-the-art set-based face recognition works on IJB-B and IJB-C 1:1 verification protocols. }
	\label{tab:SOTA}
	\vspace{-0.5cm}
\end{table*}
\noindent\textbf{Complementary with different face quality scores.}
The proposed QA-GMP enables the original GMP to be aware of both image frequency and quality. To demonstrate the generalization ability, we combine QA-GMP with three kinds of quality scores on the aggregation of VGGFace2 features.  
FaceQNet \cite{hernandez2019faceqnet} is a face quality assessment model to estimate the quality scores. NAN \cite{yang2017neural} and MCN \cite{xie2018multicolumn} are two elaborate attention blocks.
As shown in Table~\ref{tab:face_quality}, 
QA-GMP consistently improves the performance with these scores, which demonstrates its effectiveness and generalization ability on different quality scores.

\begin{table}[t]
	\centering
	\begin{tabular}{l|c|l|c}
		\toprule
		\textbf{Method}&Accuracy (\%) & \textbf{Method} & Accuracy (\%) \\
		\hline
		Eigen-PEP \cite{li2014eigen} & 84.8& DeepFace \cite{taigman2014deepface}& 91.4\\ 
		FaceNet \cite{schroff2015facenet}& 95.52 & DAN \cite{rao2017learning}& 94.28 \\
		DeepID2+ \cite{sun2015deeply} & 93.20 & QAN \cite{liu2017quality}&96.17\\
		NAN \cite{yang2017neural} & 95.72 & REAN \cite{gong2019recurrent} & 96.60\\
		Liu \emph{et al.} \cite{liu2019feature} & 96.21 & C-FAN \cite{gong2019video}& 96.50\\ 
		
		CosFace \cite{wang2018cosface} & 97.65 & ArcFace \cite{deng2019arcface}& 98.02\\ 
		\hline
		ArcFace-C \cite{deng2019arcface} & 96.66 & + GMP (T) & 96.82 \\
		+ Q-shift (E)& 97.00&+ S-sim (E)& 96.84 \\
		+ GMP(E)& 96.88 &+ QA-GMP(E)& \textbf{97.04}\\
		\bottomrule
	\end{tabular}
	\caption{Video face verification performance on YTF dataset.}
	\label{tab:YTF}
	\vspace{-0.5cm}
\end{table}
\subsection{Comparison with State-of-the-Arts} \label{sec:sota}
We compare with the state-of-the-art works on IJB-B and IJB-C datasets in Table~\ref{tab:SOTA}. The attention mechanism \cite{yang2017neural,xie2018multicolumn,xie2020inducing},  aggregation method \cite{zhong2018ghostvlad} and hard-pair mining \cite{xie2018comparator} strategy are widely studied on the VGGFace2 backbone. The proposed method is based on 
the GMP-based training and QA-GMP aggregation which usually perform best from the former comparisons. 
It can be seen that the proposed burstiness suppression methods on training and evaluation stage surpass the state-of-the-art works on both datasets with the VGGFace2 and ArcFace backbones.
Note that few of existing works report the 1:N identification performance, so we give the 1:N identification performance in the supplementary.

\noindent \textbf{Youtube Face.}  For the evaluation of YTF dataset, most of existing works adopt exhaustive (all-to-all) matching on the verification of set pairs and the maximal matching score is used as the pair similarity. This practice brings large storage and computing overhead, which is not scalable to large-scale recognition. So we adopt sum-aggregation and re-evaluate the ArcFace model on YTF with compact set representations (ArcFace-C). We then apply the GMP (T) model and four burst-aware aggregation methods for evaluation. 
As shown in Table~\ref{tab:YTF}, burst-aware training and aggregation improve  ArcFace performance. Among the four aggregation methods, Q-shift and QA-GMP get relatively better results.

\noindent \textbf{IJBC-BS.} 
As described in Section~\ref{sec:ijbcbs}, IJBC-BS protocol evaluates the performance of bursty sets of IJB-C. 
We present the performance of different models on IJBC-BS in Table~\ref{tab:IJBBS}. It can be seen that all the models suffer from performance drops and the drops of original model or vanilla training are most evident. The declines in our  GMP-based training (T) and QA-GMP (E)  are relatively not so severe. 
These observations demonstrate the hardness to recognize bursty face sets in SFR task, and the proposed burst-aware training and evaluation strategies can suppress the burstiness and improve the recognition performance.

\begin{table}[t]
	\centering
	\begin{tabular}{l|c|c|c}
		\toprule
		\multirow{2}*{\textbf{Method}} & \multicolumn{3}{c}{\textbf{IJBC-BS 1:1 TAR}}\\
		\cline{2-4} 
		~ & FAR=1e-6 & FAR=1e-5 &FAR=1e-4  \\
		\hline
		VGGFace2 \cite{cao2018vggface2} & 47.47  & 58.05 ($\downarrow16.66$) & 72.58   \\
		+ Vanilla (T)  & 49.15  & 60.95 ($\downarrow20.68$) & 74.12 \\ 
		+ \textbf{GMP (T)} & \textbf{54.35} & 67.72 ($\downarrow13.71$) &79.51   \\
		+ \textbf{GMP (T)} + \textbf{QA-GMP (E)} & 54.20 & \textbf{68.64 ($\downarrow\textbf{13.45}$)}& \textbf{79.91}  \\
		\hline
		ArcFace \cite{deng2019arcface} &  69.25 & 85.59 ($\downarrow7.56$) &91.58 \\
		+ Vanilla (T)  &73.45 & 87.82 ($\downarrow6.02$)& 92.07 \\
		+ \textbf{GMP (T)}&  74.41 & 88.13 ($\downarrow5.93$)&92.09 \\  
		+ \textbf{GMP (T)} + \textbf{QA-GMP (E)}& \textbf{74.93}& \textbf{88.32} ($\downarrow\textbf{5.74}$) & \textbf{92.13}\\
		\bottomrule
	\end{tabular}
	\caption{Ablation of burst suppression on IJBC-BS protocol.} 
	\label{tab:IJBBS}
	\vspace{-0.5cm}
\end{table}

\subsection{Qualitative Results} 

\begin{figure}[t]
	\centerline{
		\subcaptionbox{\small Vanilla \emph{vs.} GMP }{
			\includegraphics[scale = 0.29]{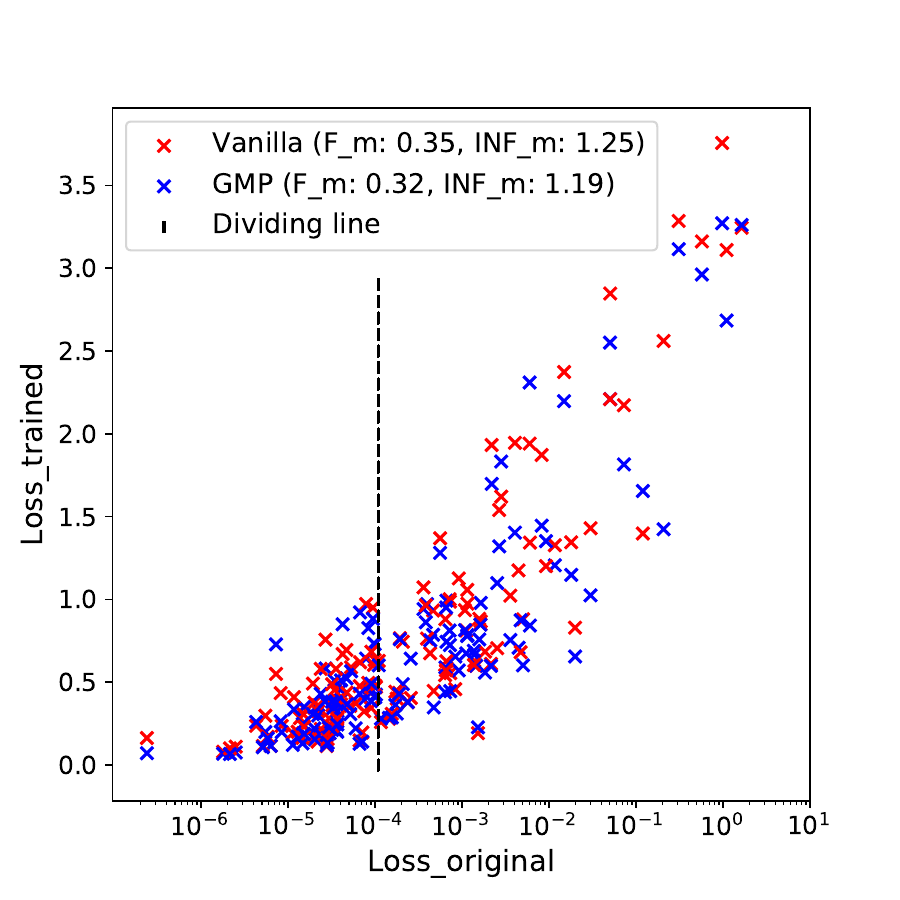}
		}\hspace{-0.5cm}
		\subcaptionbox{\small Vanilla \emph{vs.} S-sim }{
			\includegraphics[scale = 0.29]{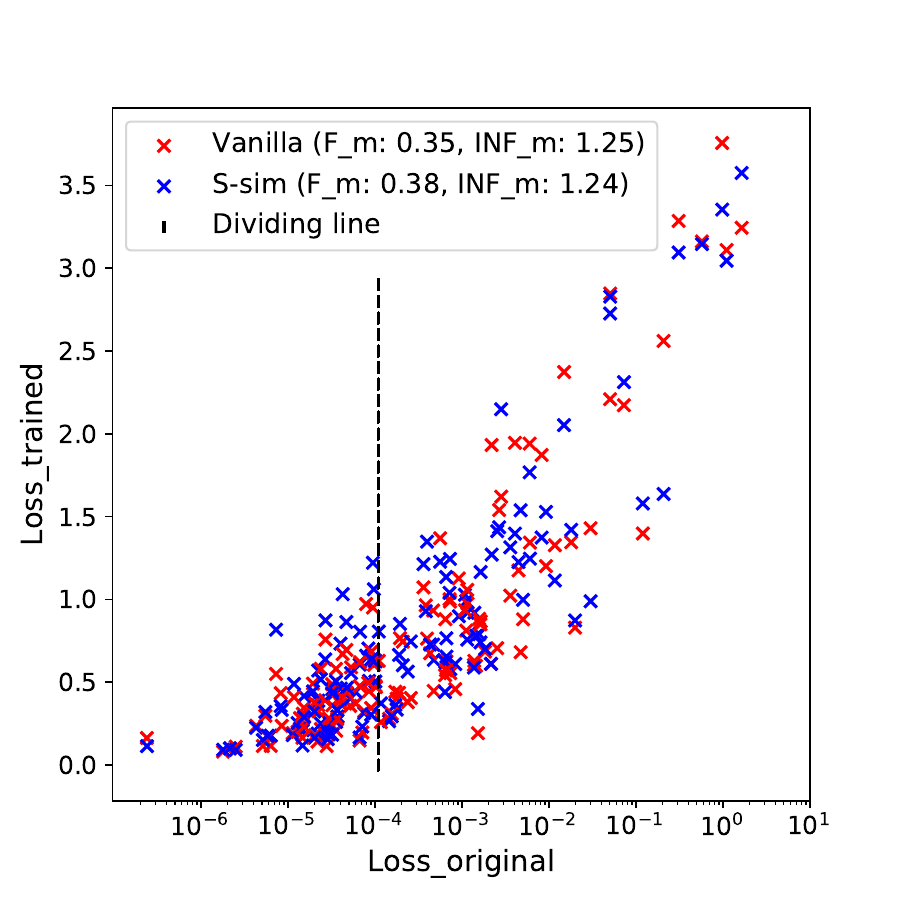}
		}\hspace{-0.2cm}
		\subcaptionbox{\small Aggregation scores }{
			\includegraphics[scale = 0.24]{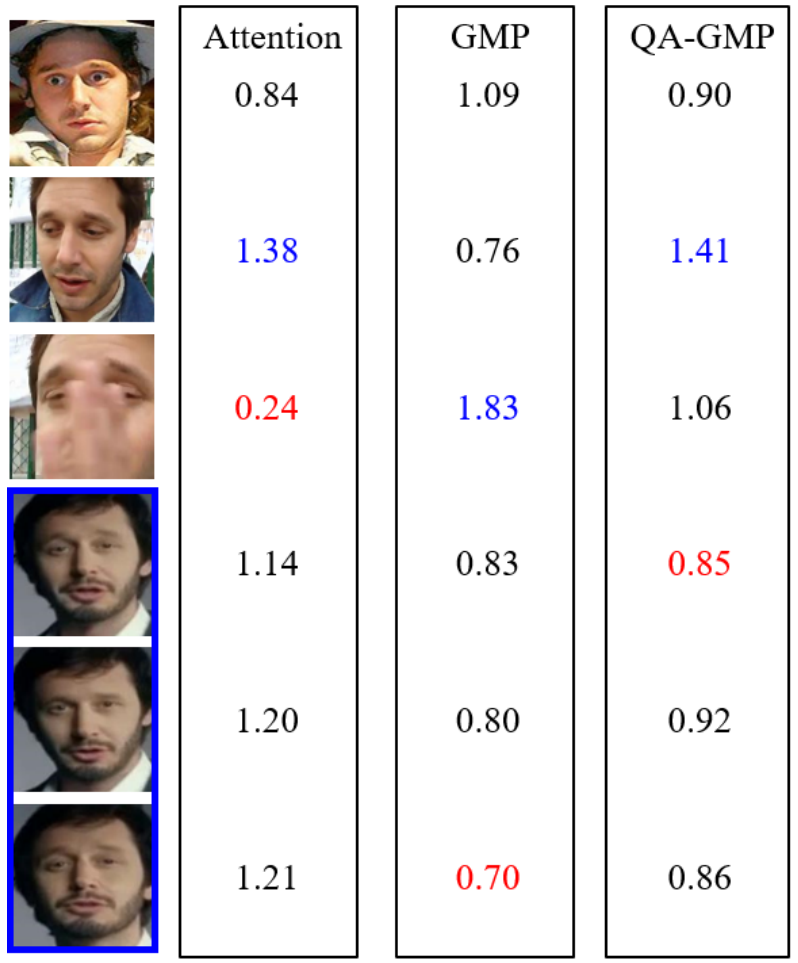}
		}
	}
	\caption{Illustration of the sample losses before and after training with three strategies in a face set. The mean trained loss of frequent faces (F\_m) and infrequent faces (INF\_m) are plotted. The dividing line divides the frequent (left) and infrequent (middle) samples with same cardinality. (Right) Illustration of the aggregation scores, which are normalized for comparison. }
	\label{fig:quali1}
\end{figure}

To reveal the effectiveness of burst-aware training, we separately train the VGGFace2 model with vanilla sampling, S-sim and GMP-based sampling. 
The corresponding three best-performed models are adopted to output the sample losses in a training face set. 
As shown in Figure~\ref{fig:quali1} (a, b), the original losses and the losses after training of all the samples are plotted. Based on the experience learned from Figure~\ref{fig:burst_train} that frequent faces usually get lower losses, half number of the faces are supposed to be frequent according to their lower original losses, and the remaining faces are supposed to be infrequent. 

Compared to vanilla sampling, both the frequent faces with lower original losses and infrequent faces with higher original losses are better learned with lower trained losses (lower F\_m and INF\_m)  when training with GMP samplings. 
While for S-sim, the infrequent faces are slightly well-learned (Lower INF\_m) than vanilla sampling by sacrificing the fitting of frequent faces (Higher F\_m).

The predicted attention, GMP and QA-GMP scores are illustrated in Figure~\ref{fig:quali1} (right). We can observe that the attention scores are usually quality-aware, while the GMP scores are frequency-aware. The third face with low quality gets lowest quality score but highest GMP score. 
QA-GMP takes both the feature quality and frequency into consideration and gives higher scores for the high-quality and low-frequency faces. Similar observations can be found in Figure~\ref{fig:burst_score}.

\section{Conclusion}
In this paper, we give an exhaustive study on the burstiness phenomenon in set-based face recognition task. We propose three strategies to detect the bursty faces in sets and separately apply them in the training stage with burst-aware instance sampling and in the evaluation stage with burst-aware aggregation. Moreover, we propose the QA-GMP to combine GMP with face quality attention, which enables the original GMP to be robust to the low-quality face features. Extensive experiments demonstrate the effectiveness of burstiness suppression with proposed strategies, which get new state-of-the-art results on the SFR benchmarks. We expect the contributions and observations to inspire other set-based recognition tasks, such as point cloud recognition and video recognition.

\clearpage
\bibliographystyle{splncs04}
\bibliography{burst}
\end{document}